\documentclass[conference]{IEEEtran}
\IEEEoverridecommandlockouts
\usepackage{cite}
\usepackage{amsmath,amssymb,amsfonts}
\usepackage{algorithmic}
\usepackage{graphicx}
\usepackage{textcomp}
\usepackage{xcolor}
\usepackage{balance}
\usepackage{multirow}
\usepackage{floatrow, subcaption}
\DeclareFloatVCode{somespace}{\vspace{1.0\baselineskip}}
\usepackage{amssymb}
\usepackage[shortcuts, acronym]{glossaries}
\makeglossaries
\newacronym{apcer}{APCER}{Attack Presentation Classification Error Rate}
\newacronym{bpcer}{BPCER}{Bonafide Presentation Classification Error Rate}
\newacronym{hter}{HTER}{Half Total Error Rate}

\def\BibTeX{{\rm B\kern-.05em{\sc i\kern-.025em b}\kern-.08em
    T\kern-.1667em\lower.7ex\hbox{E}\kern-.125emX}}
 
\newcommand{\etal}{\textit{et al.}}

\begin{document}

\title{Demographic Bias in Presentation Attack Detection of Iris Recognition Systems\\
\thanks{This research work has been funded by the German Federal Ministry of Education and Research and the Hessen State Ministry for Higher Education, Research and the Arts within their joint support of the National Research Center for Applied Cybersecurity ATHENE.}
}

\author{Meiling Fang$^{1,2}$, Naser Damer$^{1,2}$, Florian Kirchbuchner$^{1,2}$, Arjan Kuijper$^{1,2}$\\
$^{1}$Fraunhofer Institute for Computer Graphics Research IGD,
Darmstadt, Germany\\
$^{2}$Mathematical and Applied Visual Computing, TU Darmstadt,
Darmstadt, Germanyjo\\
Email: {meiling.fang@igd.fraunhofer.de}
}

\maketitle

\begin{abstract}
With the widespread use of biometric systems, the demographic bias problem raises more attention. Although many studies addressed bias issues in biometric verification, there are no works that analyze the bias in presentation attack detection (PAD) decisions.
Hence, we investigate and analyze the demographic bias in iris PAD algorithms in this paper.
To enable a clear discussion, we adapt the notions of differential performance and differential outcome to the PAD problem.
We study the bias in iris PAD using three baselines (hand-crafted, transfer-learning, and training from scratch) using the NDCLD-2013 \cite{ndcld13} database. The experimental results point out that female users will be significantly less protected by the PAD, in comparison to males.

\end{abstract}


\vspace{-1mm}
\section{Introduction}
\label{sec:intro}
\vspace{-1mm}
Biometric recognition systems like face, iris, or fingerprint have gained wide deployment within various application fields. This has raised two significant questions. One is concerned with the security of recognition systems against presentation attacks. The other is related to fairness, i.e., if the system performs the same for every different demographic group.

General presentation attacks include printed images, replaying videos, synthetic images. Iris presentation attack also contains textured contact lenses. Several researchers have already addressed these PAD problem, e.g., face PAD \cite{DBLP:journals/ivc/JiaGXW20, DBLP:journals/tifs/GeorgeMGNAM20, DBLP:conf/cvpr/ShaoLLY19} and iris PAD \cite{DBLP:journals/csur/CzajkaB18}.

Demographic bias in face recognition was discussed by \cite{howard2019, KlareBKBJ12, DBLP:journals/corr/abs-2002-03592}. Prior works on this issue indicated that face recognition systems obtained inherently a lower performance for certain demographic groups \cite{howard2019, KlareBKBJ12}. For example, females are more challenging to recognize compared to males \cite{KlareBKBJ12}. However, no previous works address demographic bias in PAD decisions.
As iris recognition has become increasingly popular, e.g., iris recognition technology is deployed on consumer smartphones, we address the demographic bias issues in iris PAD systems in this work. 

Howard \etal \cite{howard2019} defined four recognition related terms to facilitate demographic bias research. We adapt and reuse two of these terms to investigate the demographic bias in PAD systems: 1) \textbf{Differential Performance} represents a difference in the bona fide or attack decision distribution between specific demographic groups independent of any decision threshold, 2) \textbf{Differential Outcome} describes a difference in \gls{apcer} or \gls{bpcer} rates between different demographic groups relative to a decision threshold. We utilize these terms to demonstrate our experiments and examine the bias in Sec.\ref{sec:experiments} and Sec.\ref{sec:results}. 

This work presents the first analysis of demographic bias in PAD systems (of any modality). Specifically, we study gender demographic bias in both hand-crafted and deep learning-based iris PAD algorithms. This is done under the variation of the training data bias scenarios. We point out that the PAD decisions tend to be significantly less accurate for female subjects in most of the experimental setups.


\vspace{-1mm}

\section{Related Work}
\vspace{-1mm}
\label{sec:related-works}
\paragraph{Bias in Biometric Systems}
Danks \etal \cite{DBLP:conf/ijcai/DanksL17} defined the five types of algorithmic biases: 1) training data bias, 2) algorithmic focus bias, 3) algorithmic processing bias, 4) transfer context bias, 5) interpretation bias. The study of algorithmic bias in autonomous systems has caused researchers to think about bias in biometric systems. Despite the improved performance of biometric recognition systems such as fingerprint, iris or face, the fairness of the recognition systems is still questionable. The current processes in biometric systems include (among others) recognition, classification and presentation attack detection. Biometric recognition is a technology that can identify or verify a person by using biometric information \cite{ISO_recog, ISO_recog2}.
For example, facial or iris recognition systems compute a similarity score between subjects from images or a sequence of video frames. The classification here refers to assigning demographics or other labels to biometric samples \cite{DBLP:journals/tifs/DantchevaER16}. Presentation attack refers to using a presentation to fool the system in an effort to induce an incorrect decision such as using printed images or display videos. In contrast, presentation attack detection systems are expected to automatically determine such attacks \cite{ISO301071, DBLP:series/acvpr/978-3-319-92626-1}. 

Drozdowski \etal \cite{bias_survey} summarized an overview of the topic of algorithmic bias in the context of biometrics and also contributed a comprehensive survey of the existing literature on biometric bias estimation and mitigation. According to recent studies, biometric recognition and classification systems do not perform equally well in different demographic groups. Several face recognition relevant studies indicated that the female, Black, and younger ($18 - 30 y.o. $) cohorts are inherently more difficult to recognize \cite{howard2019, KlareBKBJ12}. Such phenomena that recognition systems have a lower performance for specific demographic groups is denoted as demographic bias. The study of demographic bias provides a chance to find the recognition system problems and mitigate the biases. 

\paragraph{Bias in PAD systems}
Many studies have already pointed demographic bias in face or fingerprint recognition systems \cite{DBLP:journals/corr/abs-1812-00194, howard2019, KlareBKBJ12, DBLP:journals/corr/abs-1808-05508} and obtained similar conclusions. Moreover, there is one study that experimentally related the bias to biometric attacks, namely the face morphing attack \cite{DBLP:conf/icb/Vicente-GarciaW19}. Garcia \etal \cite{DBLP:conf/icb/Vicente-GarciaW19} suggested that morphing attacks are much more successful for Asian females. However, the authors did not study bias in attack detection performance. Since iris biometric features are deployed in many security scenarios, the decision bias in the processes involved must be analyzed and minimized. Unfortunately, no previous works experimentally addressed the demographic bias in PAD systems in general, or the iris PAD specifically.
The primary difficulty is limited data resources. Czajka \etal \cite{DBLP:journals/csur/CzajkaB18} summarized iris PAD databases. Parts of the databases cannot be shared due to information protection, others were collected many years ago and are unrealistically simple. The rest of the databases contained almost no demographic information. In Sec.\ref{sec:experiments}, we find that only one database, NDCLD-2013 \cite{ndcld13}, can be used to investigate the demographic bias. Even this database also has an imbalanced number of attack samples of males and females.
Despite the limited database availability, the demographic bias in iris PAD is an open issue and is of great interest. With the widespread use of iris recognition technology, we believe the bias problem needs to be analyzed to enable future mitigation efforts.

\section{Iris Presentation Attack Detection}
\label{sec:PAD}
Iris recognition systems are susceptible to the presentation attacks, like printed iris images, replaying videos, or contact lenses. As a result, numerous iris PAD solutions have been proposed to target iris vulnerability issues \cite{DBLP:conf/cvpr/HoffmanSR18, Fang20}. Czajka \etal \cite{DBLP:journals/csur/CzajkaB18} summarized the state-of-the-art iris PAD algorithms in detecting different categories of attack. Various iris PAD frameworks are grouped into hand-crafted and deep-learning-based methods. Hand-crafted features such as Local Binary Pattern (LBP) \cite{lbp14, wlbp10, emp_eval_19}, Binarized statistical image features (BSIF) as well as their variations have been the research focal points until 2015 and made a remarkable contribution to iris PAD problems. After that, with the rapid development and application of deep learning in multiple domains, especially, computer vision field, neural-network-based PAD methods showed up out of nowhere. However, the deeper the network structure e.g., VGG16 \cite{vgg16}, the higher the computational needs. Such algorithms with high computational requirements are hard to deploy in mobile devices. Therefore, Howard \etal \cite{mobilenet} released two new MobileNet models: MobileNetV3-Large and MobileNetV3-Small to target high and low resource use cases. We utilize the MobileNetV3-Small structure to be one of our PAD baselines. 

To investigate the effects of demographic bias on different types of iris PAD systems. These systems are based on 1) hand-crafted features, 2) transfer learning of a general-purpose pre-trained network, 3) a trained from scratch network with computationally-efficient. We use the following baselines to examine the differential performance and differential outcome: 
\begin{enumerate}
    \item \textbf{LBP + SVM:}
        LBP \cite{lbp14} has been proposed to address the texture-based classification tasks. We adapt the LBP feature in conjunction with a Support Vector Machine (SVM) \cite{Cortes95support-vectornetworks} using a linear kernel to detect the presentation attacks. The threshold, which determines the label of the iris image, is chosen based on the development subset. Finally, the trained SVM model with a specific threshold is used to make a decision in the testing set, whether the iris image is bona fide or attack.
    \item \textbf{VGG-16 + PCA + SVM:}
        VGG-16 \cite{vgg16} is a convolutional neural network for image classification. Considering that VGG-16 is one of the most widely used feature extractors, we utilize the pre-trained VGG-16 to encode iris features. After extraction of VGG-16 features, a linear dimensionality reduction technique, Principal Component analysis (PCA), is used to project the extracted feature from each iris image with $7 \times 7 \times 512$ dimension into a $128$ dimensional sub-space. Similar to LBP, those lower-dimensional features are fed to SVM \cite{Cortes95support-vectornetworks} with a linear kernel and make a prediction (bona fide or attack) based on the threshold choosing from development set.
    \item \textbf{MobileNetV3-Small:}
         We use MobileNetV3-Small \cite{mobilenet} as a network structure to train from scratch. The contrast-enhanced full iris images in the training set are fed to the MobileNetV3-Small model. Again, the threshold is picked up to the development set and used to predict the label of iris image (bona fide and attack). In this stage, we use the \textit{early stopping criteria} to stop training once there is no improvement in 5 epochs. The batch size is set to 16, limited by the GPU memory.
\end{enumerate}
\vspace{-1mm}
\section{Experimental setup}
\label{sec:experiments}
\begin{figure}[!ht]
    \centering
    \includegraphics[width=0.9\linewidth]{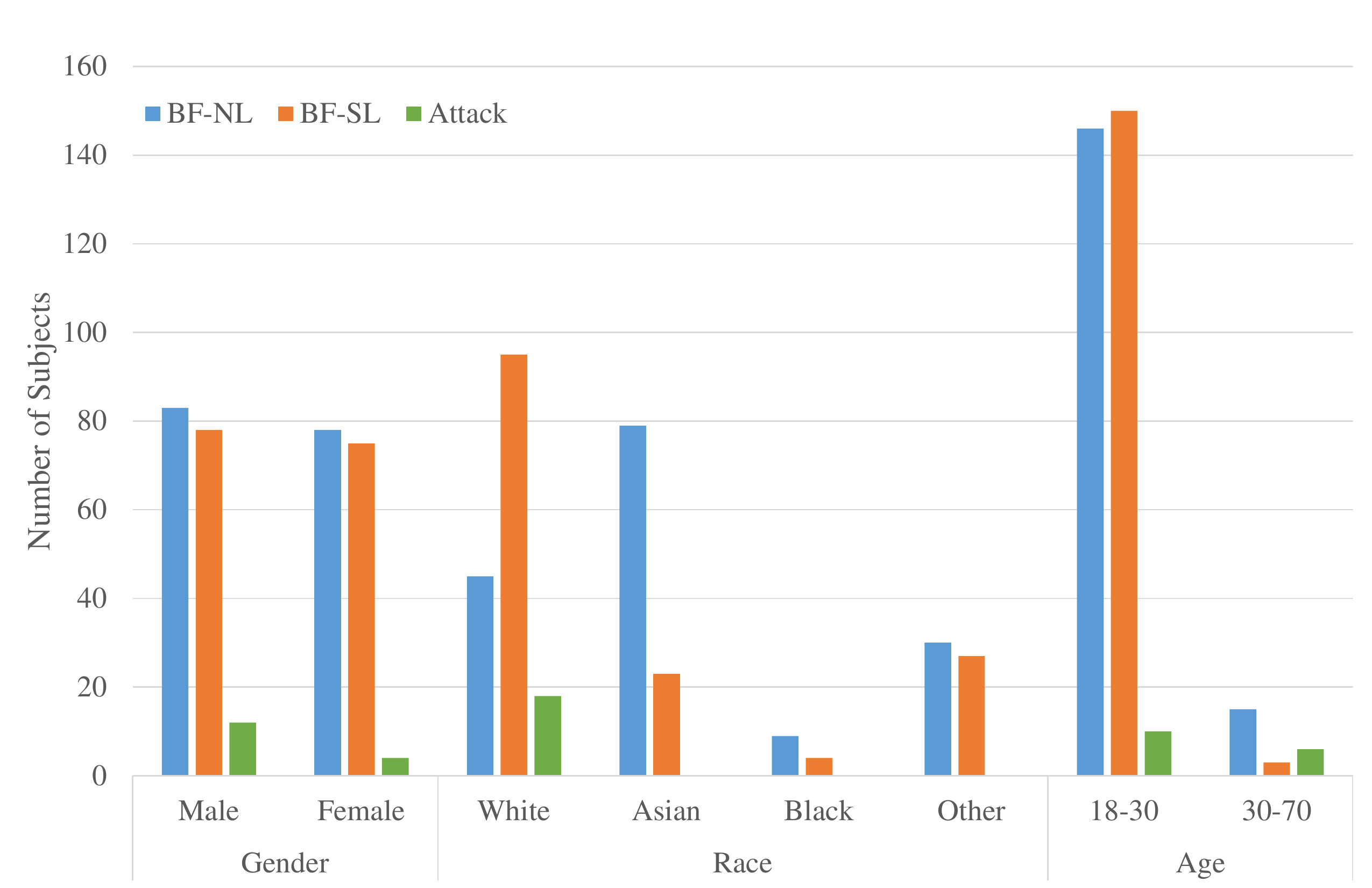}
    \vspace{-2mm}
    \caption{Subject-specific distribution of the demographic variables by different types of representation in database NDCLD-2013 \cite{ndcld13}. BF-NL is the bona fide iris image, BF-SL is the iris image with the soft lenses, which is also reported as bona fide, Attack is the iris image with textured lenses. }
    \label{fig:db_dist}
\end{figure}

\vspace{-1mm}
In this experiment, we use the NDCLD-2013 database \cite{ndcld13} including textured contact lens attack collected by the University of Notre Dame. The detailed iris image capture information has been published in \cite{ndcld13}. Briefly, this database contained $5100$ images of $330$ subjects, which acquired with a $LG4000$ and an IrisGuard $AD100$ iris capture device. At the time of the acquisition, Doyle \etal \cite{ndcld13} have divided the database into training and testing subsets, which was guaranteed to have an even distribution of genders and ethnicity. Nonetheless, we demonstrate experiments only in one database and only on the gender cohort. The reasons are:
\begin{enumerate}
    \item The number of the public iris PAD database is much less than the face and iris recognition databases. Due to the awareness of privacy security, demographic information is more difficult to collect. To our knowledge, only NDCLD-2013 \cite{ndcld13} and NDCLD-2015 \cite{ndcld15} database had demographic information. While the NDCLD-2013 included gender, race and age categories, the NDCLD-2015 had gender and eye color information.
    \item There are only two female subjects with attack samples in the NDCLD-2015 database. Similarly, the eye color was largely unbalanced in this database. Therefore, we have to disregard the NDCLD-2015 database, though we believe that the impact of eye color bias in iris PAD systems is very interesting and worth the investigation.
    \item As shown in Fig. \ref{fig:db_dist}, blue (BF-NL) denotes the number of subjects wearing no contact lens, green (BF-SL) is the number of subjects wearing the soft lenses and green (attack) is the number of subjects wearing the textured lenses. It should be noticed that race/ethnicity and age categories are unbalanced: 1) no attack samples in Black and Asian categories, 2) unbalance age groups ($18 - 30 y.o.$), and ($30 - 70 y.o.$). Therefore, for statistical significance, we only analyze the demographic bias over gender.
\end{enumerate}

\vspace{-2mm}
\vspace{-2mm}

Given the limited data resources, we are only able to study gender demographics in iris PAD systems. Two sub-database LG4000 and AD100 are combined into one database. Train-test-split is defined in \cite{ndcld13}. A summary of the training and testing samples for each type of representation can be found in Tab.\ref{tab:dataset_setup}. Additionally, we split the training subset into identity-disjoint training ($80\%$) and development ($20\%$) sets. In this study, soft lens iris images are used as bona fide samples.


\begin{table}[!ht]
\footnotesize
\centering
\scalebox{0.83}{
\begin{tabular}{|c|c|c|c|c|c|c|}
\hline
Gender  & \multicolumn{3}{c|}{Male} & \multicolumn{3}{c|}{Female} \\ \hline
Type & BF-NL & BF-SL & Attack & BF-NL & BF-SL & Attack   \\ \hline
\# Training & 600 & 600 & 920 & 600 & 600 & 280      \\ \hline
\# Testing  & 250 & 250 & 370 & 250 & 250 & 130      \\ \hline
\end{tabular}
}.\vspace{-2mm}
\caption{Number of images used for training and testing per gender. Training and test sets are subject-disjoint. A total of $5100$ iris images are used in this study.}
\label{tab:dataset_setup}
\end{table}
\vspace{-2mm}

For gender demographics, two experiments are conducted. Because subjects cannot be balanced in 5-fold, we repeat each experiment 5 times to replace the 5-fold cross-validation and both experiments are subject-disjoint. The first experiment measures the iris PAD performance on males versus females. Specifically, the training and development set contains iris images from both male and female subjects. Then, we separately evaluate males and females set. The result from the first experiment can be found in Tab.\ref{tab:mixed_res}. In addition, we report the attack and bona fide decision distribution in the gender cohort from three iris PAD baselines to analyze the differential performance mentioned in Sec.\ref{sec:intro} (see Fig.\ref{fig:lbp_dist}, Fig.\ref{fig:mobilenet_dist}, Fig.\ref{fig:vgg_dist}). The second experiment investigates the influence of the training subset on iris PAD performance, which means that we use iris images from one gender as a training and development set. For example, we train a model by using only male data from the training set and compute a threshold by fixing APCER at $0.01\%$ in the male development set. Then, this model and threshold are applied separately to males and females testing set. Subsequently, we are able to analyze the outcome differences in the scenarios where one gender is exclusively used for training and threshold assignment.
The detailed results of three PAD methods from this experiment can be found in Tab.\ref{tab:lbp}, Tab.\ref{tab:mobilenet} and Tab.\ref{tab:vgg}.

The following metrics are used to measure the iris PAD algorithm performance:
\begin{itemize}
	\item \textbf{Attack Presentation Classification Error Rate (APCER)}: The proportion of attack images incorrectly classified as bona fide smples.
	\item \textbf{Bonafide Presentation Classification Error Rate (BPCER)}: The proportion of bonafide images incorrectly classified as attack samples. 
	\item \textbf{\gls{hter}}: corresponds to the average of BPCER and ACPER.

The \gls{apcer}, \gls{bpcer}, and \gls{hter} that we report are calculated on the same threshold, which chosen by fixing \gls{apcer} at $0.01\%$ in the development subset. The \gls{apcer} and \gls{bpcer} follows the standard definition presented in the ISO/IEC 30107-3 \cite{ISO301073}.
\end{itemize}

\section{Results and Analysis}
\label{sec:results}
In this section, we analyze the experimental results described in Sec.\ref{sec:experiments}. First, the difference in bona fide and attack PAD decision distribution between males and females, independent of any threshold, is demonstrated to measure the differential performance. Second, a difference in BPCER and APCER between males and females regarding a specific threshold is discussed to examine the differential outcome.

\begin{figure}[!ht]
    \centering
    \includegraphics[width=1.0\linewidth]{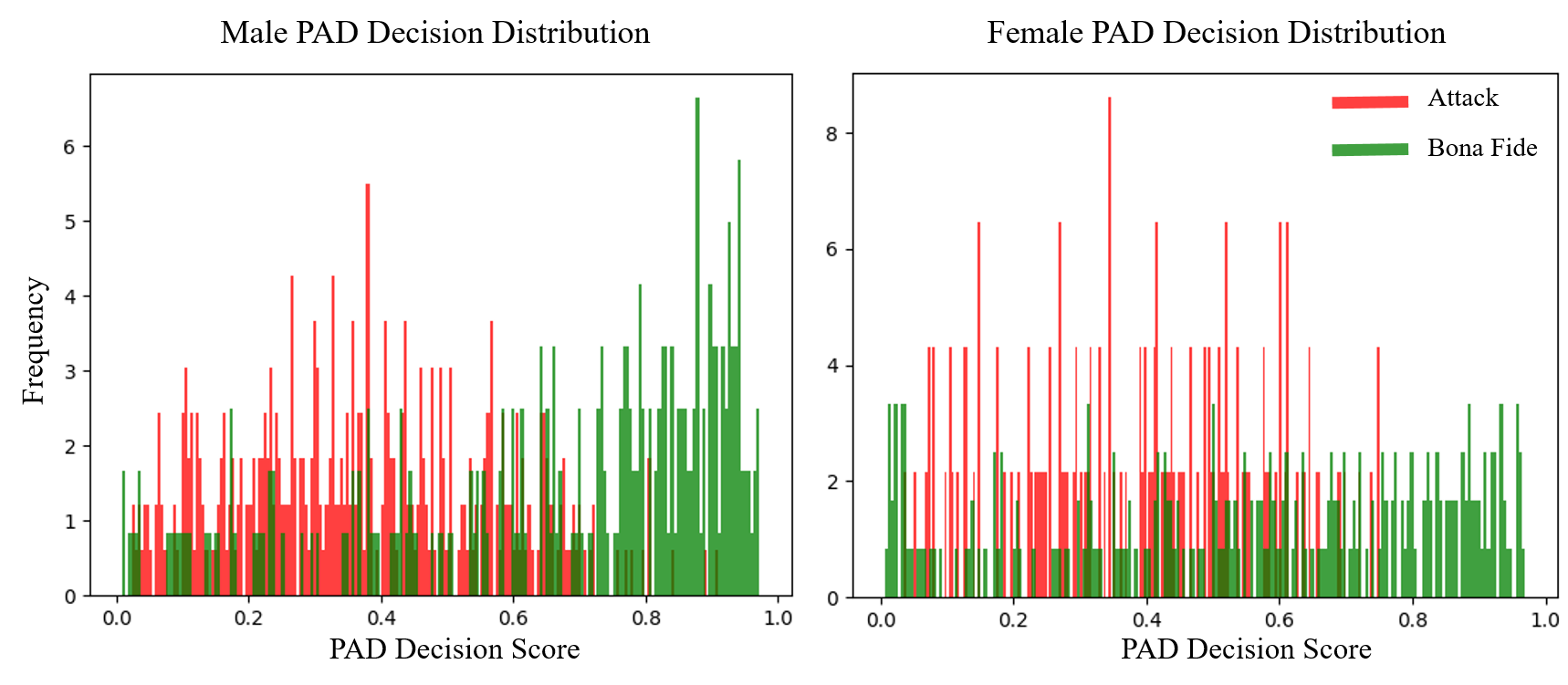}
    \vspace{-2mm}
    \caption{PAD Decision Distribution by LBP method between bona fide (green) and attack (red) on gender demographic. The female's PAD decision distribution is less separable.}
    \label{fig:lbp_dist}
\end{figure}

\begin{figure}[!ht]
    \centering
    \includegraphics[width=1.0\linewidth]{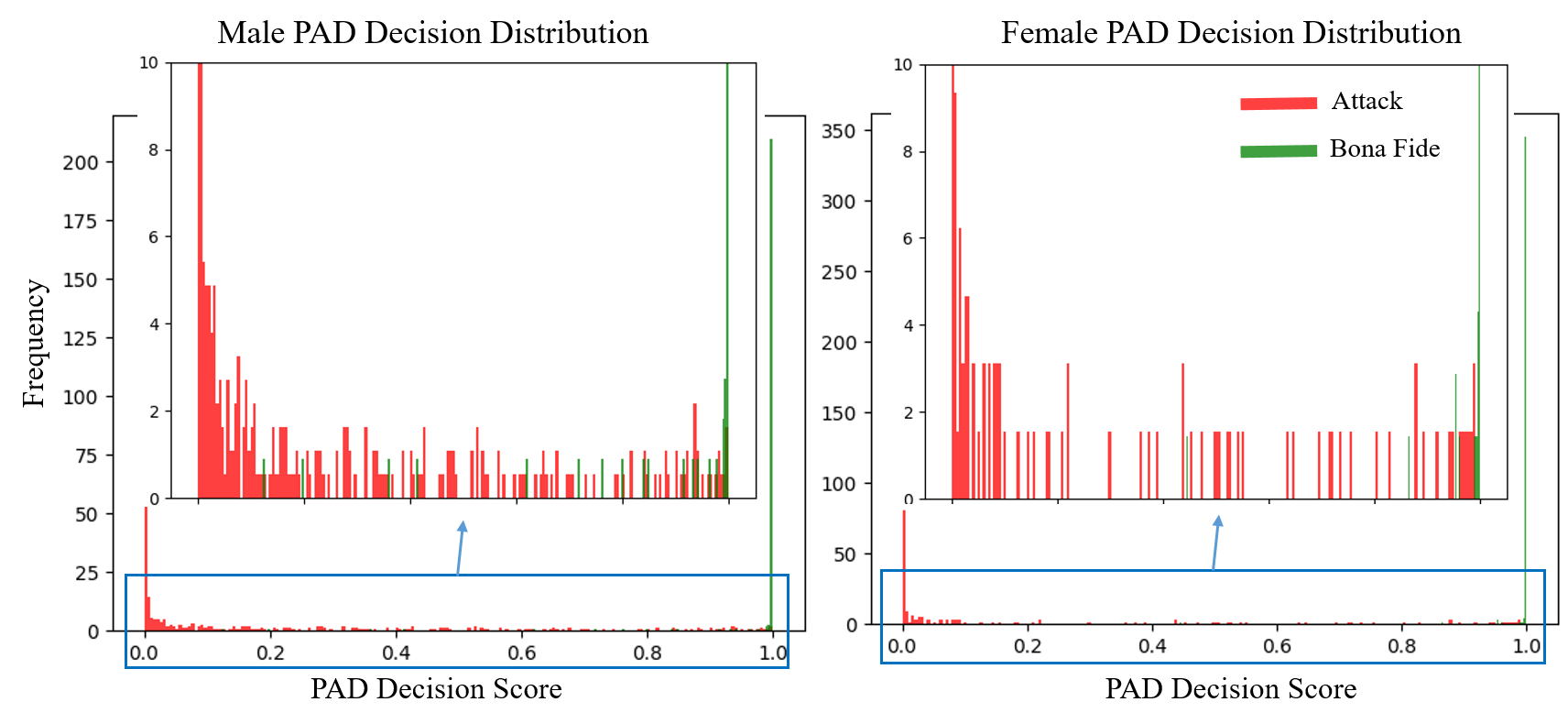}
    \vspace{-2mm}
    \caption{PAD Decision Distribution from MobileNetV3-Small method between bona fide (green) and attack (red) on gender demographic. }
    \label{fig:mobilenet_dist}
\end{figure}

\begin{figure}[!ht]
    \centering
    \includegraphics[width=1.0\linewidth]{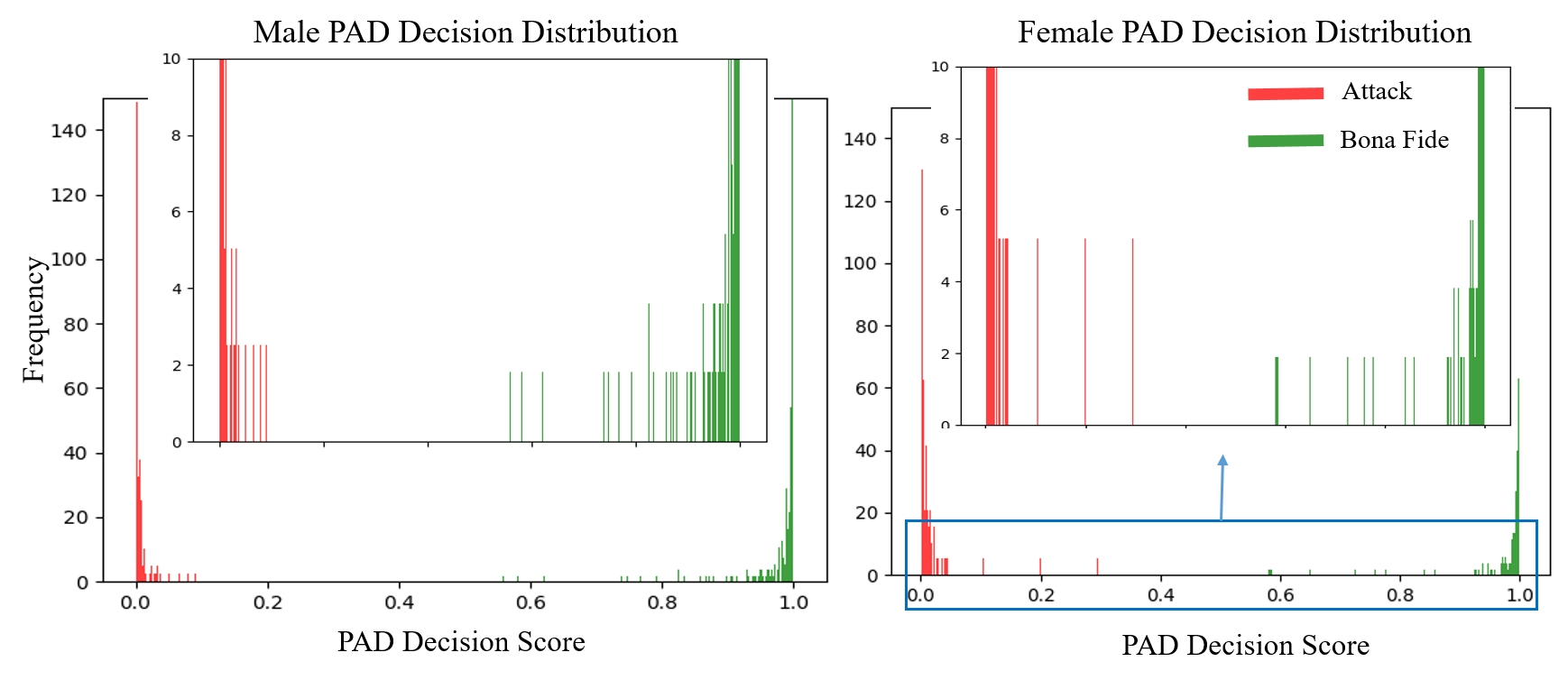}
    \vspace{-2mm}
    \caption{PAD Decision Distribution from VGG-16 method between bona fide (green) and attack (red) on gender demographic. The decision gap of female samples between bona fides and attacks are smaller than that of males.}
    \label{fig:vgg_dist}
\end{figure}

\paragraph{Differential Performance in the PAD decision distribution}
\label{ssec:diff_outcome}
The ideal distribution of PAD decision is that the decision of the attack samples is close to zero and the decision of the bona fide samples is close to one. In this subsection, we represent the PAD decision distributions between the gender cohort, which are generated by three algorithms: LBP, MobileNetV3-Small, and VGG-16 method.
First, Fig.\ref{fig:lbp_dist} shows that most bona fide decisions of males are distributed between $0.7$ to $1.0$ by LBP method. In contrast, the attack and the bona fide decisions of females are relatively difficult to separate. 
Second, deep-learning-based algorithms obtain a much better performance compared to the LBP method. As shown in Fig.\ref{fig:mobilenet_dist} and Fig.\ref{fig:vgg_dist}, the attack and bona fide decisions generated by MobileNetV3-Small and VGG-16 baselines are relatively separated. Although both gender cases achieved good results, we find out that there is still a small difference in the PAD decision distribution. By observing the zoomed region (in Fig. \ref{fig:mobilenet_dist}) in MobileNetV3-Small distribution, we notice that male decisions are denser. Moreover, the female cohort performs slightly worse than males, which is noted by the smaller gap between the attack and the bona fide distribution from VGG-16 (see Fig.\ref{fig:vgg_dist}). Together, those evidences point out that female presentation subjects are potentially more difficult to detect by iris PAD systems.

\vspace{-2mm}

\vspace{-2mm}

\vspace{-2mm}

\paragraph{Differential Outcome in APCER and BPCER}
\label{ssec:diff_outcome}
As shown in Tab.\ref{tab:mixed_res}, both hand-crafted-based and neural network-based algorithms performed worse on the females than males. This finding is consistent with the results of gender cohort researches in face recognition systems \cite{KlareBKBJ12, howard2019}. LBP method performs significantly worse on females. The BPCER of the LBP method on females ($42.60\%$) is much higher than on males ($21.85\%$) with the same number of bona fide images in the testing set. MobileNetV3-Small and VGG-16 obtain higher APCER on females, but, the same BPCER on both cohorts. Excluding the gender bias effect, one of the possible reasons is due to a small number of female attack samples in the training data as described in \cite{DBLP:conf/ijcai/DanksL17}.  

The results of the second experiment possess more information about the nature of the discrepancy. In most cases, better testing performances are achieved on the trained gender cohort when training exclusively on males or females. Also, the LBP method's performance indicates the significant error increase when training only on mixed data (males + females) than training on males/females in Tab.\ref{tab:lbp}. One possible reason is that fewer data cause the underfit or overfit of the classifier. As shown in Tab. \ref{tab:lbp}, even if the female samples have been learned, the female group performs worse than males. Hence, the LBP method shows that it is more difficult to decide if a female subject is bona fide or attack. Besides, the VGG-16 method demonstrates a much worse generalizability on the female cohort (Tab.\ref{tab:vgg}). The model, which is trained only by female samples, marks male attack samples as bona fide images. MobileNetV3-Small (Tab.\ref{tab:mobilenet}) provide the consistent generalizability of the females with the VGG-16 method. There is still a possibility that iris PAD error on females can be reduced by merely changing the distribution of the female samples. In general, we notice that different PAD solutions perform significantly better on male samples, even in some cases where females samples are exclusively used for training.

\begin{table}[h]
\centering
\scalebox{0.65}{
\begin{tabular}{|c|c|c|c|c|c|c|c|c|c|}
\hline
\multirow{2}{*}{Metric} & \multicolumn{3}{c|}{\textbf{LBP}} & \multicolumn{3}{c|}{\textbf{MobileNetV3-Small}} & \multicolumn{3}{c|}{\textbf{VGG-16}} \\ \cline{2-10} 
 & Male & Female  & Both  & Male  & Female & Both & Male & Female & Both    \\ \hline
APCER & 12.70 & 8.46 & 11.6  & 8.11 & 9.23 & 8.40 & 12.43 & 16.15 & 13.40   \\ \hline
BPCER & 31.00 & 42.6 & 36.8  & 0.20  & 0.20 & 0.20 & 0 & 0 & 0       \\ \hline
HTER & 21.85 & \textbf{25.53} & 24.2 & 4.16 & \textbf{4.72} & 4.30 & 6.22 & \textbf{8.08} & 6.7 \\ \hline
\end{tabular}
}
\vspace{-2mm}
\caption{Results of three iris PAD methods. The training and development set comprise of both males and females samples. The threshold is chosen by fixing APCER at $0.01\%$ in development set. Then, the evaluation is applied on males, females and mixed testing set separately. The bold number denotes the worst results. All HTER results indicate that the female samples are harder to classify.}
\label{tab:mixed_res}
\end{table}
\vspace{-2mm}

\begin{table}[!ht]
    \centering
    \scalebox{0.8}{
        \begin{tabular}{|c|c|c|c|c|c|c|} \hline
           & \multicolumn{6}{c|}{\textbf{LBP}} \\ \hline
           Train & \multicolumn{3}{c|}{Male} & \multicolumn{3}{c|}{Female} \\ \hline
           Test  & Male & Female & Both & Male & Female & Both \\ \hline
           APCER & 14.59 & 16.92 & 15.2 & 14.32 & 16.15 & 14.78 \\ \hline
           BPCER & 32.00 & 43.80 & 39.70 & 60.60 & 67.00 & 63.80 \\ \hline
           HTER  & 23.29 & \textbf{30.36} & 27.45 & 37.46 & \textbf{41.58} & 38.29 \\ \hline
        \end{tabular}
    }
    \vspace{-2mm}
    \caption{Results of the LBP methods. The training set and development set contain iris images of only one gender. The evaluation is performed on males, females and mixed gender, separately. Training on female cohort achieves much worse HTER than training on males. One possible reason is that Female cohort has less subjects.}
    \label{tab:lbp}
\end{table}
\vspace{-3mm}

\begin{table}[!ht]
    \centering
    \scalebox{0.8}{
        \begin{tabular}{|c|c|c|c|c|c|c|} \hline
           & \multicolumn{6}{c|}{\textbf{MobileNetV3-Small}} \\ \hline
           Train & \multicolumn{3}{c|}{Male} & \multicolumn{3}{c|}{Female} \\ \hline
           Test  & Male & Female & Both & Male & Female & Both \\ \hline
           APCER & 2.16 & 7.69 & 3.60 & 44.32 & 10.76 & 35.6 \\ \hline
           BPCER & 2.00 & 0 & 1.00 & 6.00 & 7.60 & 6.80  \\ \hline
           HTER  & 2.08 & \textbf{3.85} & 2.30 & \textbf{25.16} & 9.18 & 21.20 \\ \hline
        \end{tabular}
    }
    \vspace{-2mm}
\caption{Results of the MobileNetV3-Small methods. The training set and development set contain iris images of only one gender. The evaluation is performed on males, females and mixed gender, separately.}
\label{tab:mobilenet}
\end{table}
\vspace{-3mm}

\begin{table}[!ht]
    \centering
    \scalebox{0.8}{
        \begin{tabular}{|c|c|c|c|c|c|c|} \hline
           & \multicolumn{6}{c|}{\textbf{VGG-16}} \\ \hline
           Train & \multicolumn{3}{c|}{Male} & \multicolumn{3}{c|}{Female} \\ \hline
           Test  & Male & Female & Both & Male & Female & Both \\ \hline
           APCER & 18.38 & 70.77 & 29.80 & 100 & 13.08 & 75.80 \\ \hline
           BPCER & 0 & 0 & 0 & 0 & 0 & 0  \\ \hline
           HTER  & 9.19 & \textbf{35.39} & 14.90 & \textbf{50.00} & 6.54 & 37.90 \\ \hline
        \end{tabular}
    }
    \vspace{-2mm}
    \caption{Results of the VGG-16 methods. The training set and development set contain iris images of only one gender. The evaluation is performed on males, females and mixed gender, separately.}
    \label{tab:vgg}
\end{table}

\section{Conclusion}

In this paper, we address the demographic bias problem in iris PAD systems. Since no previous work discussed bias in PAD, we adapt two terms used for verification bias (differential performance and outcome) to analyze the demographic bias on PAD performance. Two kinds of experiments are designed to investigate demographic bias in iris PAD accuracy, as well as the impact of the training data bias. The results achieved by three PAD baselines showed a significant difference in the performance and outcome of the algorithms between male and female samples in most experimental setups.

\balance
\bibliographystyle{IEEEtranS}
\bibliography{EUSIPCO2020}

\end{document}